\documentclass{article}
\pdfoutput=1
\usepackage[utf8]{inputenc}
\usepackage[top=30truemm,bottom=30truemm,left=25truemm,right=25truemm]{geometry}
\usepackage[dvipdfmx]{graphicx}
\usepackage{amsmath}
\usepackage{amsfonts}
\usepackage[symbol]{footmisc}
\usepackage{amsthm}
\usepackage{fancybox}
\usepackage{color}
\usepackage{cases}
\usepackage{subfiles}

\usepackage{authblk}

\usepackage{stfloats}
\fnbelowfloat

\newtheorem{theorem}{Theorem}
\theoremstyle{definition}
\newtheorem{definition}{Definition}

\newtheorem{property}{Property}
\usepackage{todonotes}

\title{Stability-Certified Reinforcement Learning\\
via Spectral Normalization}
\author[1]{\fontsize{11.5pt}{0pt}\selectfont Ryoichi Takase\footnote{Contact Author (email: takase-aero@g.ecc.u-tokyo.ac.jp)}}
\author[2]{\fontsize{11.5pt}{0pt}\selectfont Nobuyuki Yoshikawa}
\author[2]{\fontsize{11.5pt}{0pt}\selectfont Toshisada Mariyama}
\author[1]{\fontsize{11.5pt}{0pt}\selectfont Takeshi Tsuchiya}
\affil[1]{\fontsize{10.5pt}{0pt}\selectfont Department of Aeronautics and Astronautics, The University of Tokyo}
\affil[2]{\fontsize{10.5pt}{0pt}\selectfont Information Technology R\&D Center, Mitsubishi Electric Corporation}
\date{December 2020}

\begin{document}

\maketitle

\begin{abstract}
In this article, two types of methods from different perspectives based on spectral normalization are described for ensuring the stability of the system controlled by a neural network. The first one is that the $\mathcal{L}_2$ gain of the feedback system is bounded less than 1 to satisfy the stability condition derived from the small-gain theorem. 
While explicitly including the stability condition, the first method may provide an insufficient performance on the neural network controller due to its strict stability condition. 
To overcome this difficulty, the second one is proposed, which improves the performance while ensuring the local stability with a larger region of attraction. In the second method, the stability is ensured by solving linear matrix inequalities after training the neural network controller. 
The spectral normalization proposed in this article improves the feasibility of the a-posteriori stability test by constructing tighter local sectors. The numerical experiments show that the second method provides enough performance compared with the first one while ensuring enough stability compared with the existing reinforcement learning algorithms. 
\end{abstract}

\section{Introduction}
Reinforcement learning (RL) has achieved impressive success in a variety of domains: the games of Go~\cite{silver2016mastering} and Atari~\cite{mnih2013playing,mnih2015human}. 
These successes have led to an interest in solving real-world problems such as robotics~\cite{zhu2020ingredients,jin2020offline}, automation driving~\cite{Osinski_2020,Almasi_2020}, and unmanned aerial vehicle~\cite{koch2019neuroflight,koch2019reinforcement}. While providing high performance, RL algorithms suffer from a lack of safety certificates required for safety-critical systems. 
The difficulty of the safety certificates is seen in, for example, the stability guarantees of the system controlled by a neural network. The neural network generally has nonlinear activation functions. As the number/size of the hidden layers is increased, the feedback system tends to have high nonlinearities, resulting in the fails of the stability tests for the neural network controller. 
In order to tackle the works on ensuring the stability guarantees, understanding of stability and RL algorithms are required, which suggests that the stability-certified RL is an essential topic in both fields of control theory and RL researches.

The Lyapunov theorem is fundamental for stability analysis of dynamical systems. Analysis based on the Lyapunov theorem needs to construct a Lyapunov function to ensure the stability~\cite{khalil2002nonlinear}. 
The Lyapunov stability methods have been widely used in the field of control engineering~\cite{khalil2002nonlinear} and recently well investigated for RL algorithms to ensure the stability~\cite{han2020actor}. 
As another method, the input-output stability is also used to achieve the same purpose~\cite{jin2018stability}.

Unlike these approaches, this article revisits the spectral normalization~\cite{miyato2018spectral} and shall propose two types of methods from different perspectives. 
The first one is to ensure the global stability of the feedback system based on the small-gain theorem. 
The stability condition is derived with the spectral normalization. 
However, this condition is highly conservative and may result in insufficient control performance. 
To overcome this difficulty, the second one is proposed, which improves the performance while ensuring the local stability with a larger region of attraction (ROA). In the second method, the stability is ensured by solving linear matrix inequalities (LMIs) after training the neural network controller. 
The spectral normalization improves the feasibility of the a-posteriori stability test by constructing tighter local sectors. 
The contribution of this article is summarized as follows:
\begin{itemize}
\item In the development of the first method, a relationship between the spectral normalization and the global stability of the feedback system is described from a control point of view,
\item Regarding the local stability, the second method improves the feasibility of the a-posteriori stability test and enlarges the ROA,
\item Numerical experiments show that the second method provides enough stability and gives almost the same performance level compared with the existing RL algorithms. 
\end{itemize}

The layout of this article is as follows. 
Section~\ref{Sec:RW} describes the related works on this research.
Section~\ref{Sec:Preliminaries} describes the problem formulation and the stability analysis of the feedback system. 
Section~\ref{Sec:Method} describes the stability-certified RL based on the spectral normalization.
Section~\ref{Sec:Experiments} gives the results of numerical experiments, and Section~\ref{Sec:Conclusion} ends the article with the conclusions.

The notation is fairly standard. 
The symbol $\mathbb{S}^n_{++}$ denotes the sets of $n$-by-$n$ symmetric positive definite matrices, and $M>0$ ($M<0$) means that $M$ is positive (negative) definite. 
If $v$, $w\in\mathbb{R}^n$ then the inequality $v\le w$ is element-wise, i.e. $v_i\le w_i$ for $i=1,...,n$.
The symbol $||x||_2$ denotes the 2-norm of the signal $x$, and $||x(k)||_{\mathcal{L}_2}$ denotes the total energy of a series of a discrete-time signal $\{x(k)\}^\infty_{k=0}$, i.e., $||x(k)||_{\mathcal{L}_2}:=\sqrt{\sum_{k=0}^\infty ||x(k)||^2_2}$.

\section{Related Works}
\label{Sec:RW}
This article is closely related to the works on RL with stability guarantees. 
The key points of the related works are summarized as follows. 
The first is the robust control framework to ensure the nominal/robust stability of the systems under the presence of nonlinearities and uncertainties. 
In Ref.~\cite{kretchmar2000synthesis}, integral quadratic constraints (IQCs) are used to deal with the nonlinear activation function and the variation of the weight matrices in the neural network.
The IQC framework has successfully been combined with the RL researches such as recurrent neural network~\cite{anderson2007robust}, semidefinite feasibility problems with bounded gradient~\cite{jin2018stability}, and local quadratic constraints for obtaining the ROA~\cite{yin2020stability}. 
As other examples using the robust control framework, the neural network certification algorithm~\cite{wang2019verification} and the $H_\infty$ learning control methods~\cite{morimoto2001robust,han2019h} have been proposed to obtain robustness against adversarial conditions in environments, see also~\cite{luo2014off,donti2020enforcing,zhang2020policy} for the interesting works.

The second key point is the technique to jointly learn the policy and the Lyapunov function. 
In Refs.~\cite{chang2019neural,richards2018lyapunov}, a candidate of the Lyapunov function is represented by the neural network, and then, the policy and the Lyapunov function are obtained through the training, which provides a larger ROA than that of the classical linear quadratic regulator (LQR). 
In Ref.~\cite{jin2020neural}, the barrier function is also learned to ensure that the policy must not run into unsafe regions.
These methods can be applied to a large class of nonlinear systems and have the potential for further improvement on the related works. 

This article differs from the above works in the aspect that the stability constraint is given by the spectral normalization. 
Specifically, this article first proposes the method for ensuring the global stability of the feedback system. In the first method, the $\mathcal{L}_2$ gain of the feedback system is bounded less than 1, that is, the spectral normalization is performed to satisfy the small-gain theorem. 
However, the first one may provide insufficient performance due to its conservative stability condition. 
To improve the practicality, the second method is proposed, which provides enough performance while ensuring the local stability with a larger ROA. 
To conclude this section, this article presents a trade-off between stability and performance of the spectral normalization on the neural network controller through theoretical groundwork and numerical experiments.

\section{Preliminaries}
\label{Sec:Preliminaries}
\subsection{Problem Formulation}

\begin{figure}[!!b]
\begin{center}
\includegraphics[width=5cm]{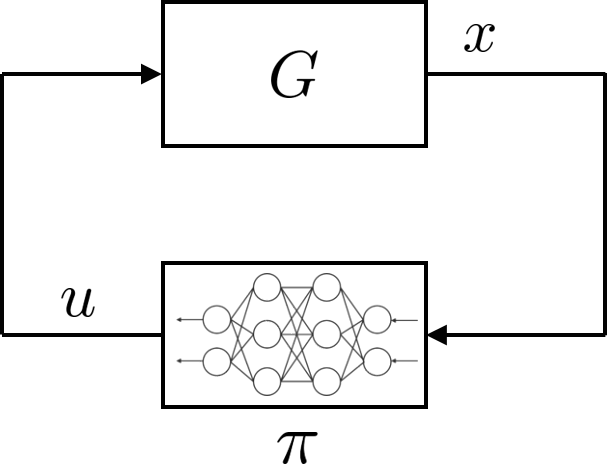}
\caption{Feedback system consisting of $G$ and $\pi$.}
\label{Fig:FeedbackNN}
\end{center}
\end{figure}

Let us consider the feedback system consisting of a plant $G$ and a controller $\pi$ as shown in Fig.~\ref{Fig:FeedbackNN}.
In this article, we assume that the plant $G$ is a discrete-time linear time-invariant (LTI) model represented as follows:
\begin{eqnarray}
\label{Eq:Plant}
x(k+1)=A_Gx(k)+B_Gu(k),
\end{eqnarray}
where $x(k)\in\mathbb{R}^{n_G}$, $u(k)\in\mathbb{R}^n_u$, $A_G\in\mathbb{R}^{n_G\times n_G}$, and $B_G\in\mathbb{R}^{n_G\times n_u}$ are the state, the input, the state matrix, and the input matrix, respectively.
The controller $\pi:\mathbb{R}^{n_G}\to\mathbb{R}^{n_u}$ is an $\ell$-layer neural network with nonlinear activation functions in hidden layers:
\begin{subequations}
\label{Eq:OrdinalNeuralController}
\begin{align}
w^0(k) &= x(k); \\\label{Eq:OrdinalNeuralControllerb}
w^i(k) &= \phi^i (W^i w^{i-1}(k)+b^i), \quad i = 1, ..., \ell; \\
u(k) &= W^{\ell+1} w^\ell(k)+b^{\ell+1},
\end{align}
\end{subequations}
where $w^i\in\mathbb{R}^{n_i}$ is the output from the $i^{th}$ layer.
Symbols $W^i \in \mathbb{R}^{n_i \times n_{i-1}}$ and $b^i\in\mathbb{R}^{n_i}$ are respectively the weight matrix and the bias of the $i^{th}$ layer in the neural network.
Symbol $\phi^i:\mathbb{R}^{n_i} \rightarrow \mathbb{R}^{n_i}$ is the activation function which is applied element-wise for a given vector $v$:
\begin{eqnarray}
\phi^i(v):=[\varphi(v_1), ..., \varphi(v_{n_i})]^T,
\end{eqnarray}
where $\varphi:\mathbb{R}\to\mathbb{R}$ is the scalar activation function, e.g., $\tanh$, ReLU, and sigmoid.
In RL, the neural network defined by Eq.~(\ref{Eq:OrdinalNeuralController}) is trained by some algorithms, e.g., trust region policy optimization (TRPO)~\cite{schulman2015trust}, proximal policy optimization (PPO)~\cite{schulman2017proximal}, and soft actor-critic (SAC)~\cite{haarnoja2018soft}. 
In this article, the controller $\pi$ is also referred to as the policy in accordance with the field of RL researches.

\subsection{Stability of Feedback Systems}
\subsubsection{Small-Gain Theorem}
 The following definition of the system stability is introduced.
 
\begin{definition}
\label{Def:L2Stability}
{\bf $\mathcal{L}_2$ Stability\\}
{\it
Consider a system whose input-output relation is represented by a mapping $H: x\to y$.
The $\mathcal{L}_2$ gain of the system is the worst-case ratio between the total output and the total input energy as follows:
\begin{eqnarray}
\label{Eq:L2Stability}
\gamma:= \sup_{x\neq0} \frac{||y||_{\mathcal{L}_2}}{||x||_{\mathcal{L}_2}}.
\end{eqnarray}
If $\gamma$ is finite, then the system is said to be finite-gain $\mathcal{L}_2$ stable.
}
\end{definition}

\begin{figure}[!!t]
\begin{center}
\includegraphics[width=6cm]{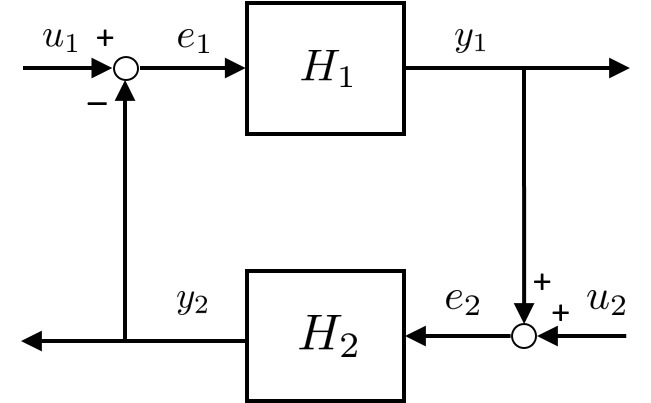}
\caption{Feedback connection.}
\label{Fig:FeedbackConnection}
\end{center}
\end{figure}

Let us consider the feedback connection as shown in Fig.~\ref{Fig:FeedbackConnection}. 
The small-gain theorem gives a sufficient condition for the feedback connection.

\begin{theorem}
{\bf Small-Gain Theorem\\}
Suppose that the systems $H_1$ and $H_2$ respectively have the finite $\mathcal{L}_2$ gains $\gamma_1$ and $\gamma_2$. 
Suppose further that the feedback system is well defined in the sense that for every pair of inputs $u_1$ and $u_2$, there exist unique signal pairs of $(e_1, y_2)$ and $(e_2, y_1)$ as shown in Fig.~\ref{Fig:FeedbackConnection}.
Then, the feedback connection is finite-gain $\mathcal{L}_2$ stable if $\gamma_1\gamma_2<1$.
\end{theorem}

\begin{proof}
See \cite{khalil2002nonlinear}.
\end{proof}

\subsubsection{Lyapunov Condition}
In this subsection, the stability analysis for the neural network controller proposed in~\cite{yin2020stability} is briefly described. By introducing local quadratic constraints (QCs) for activation functions, the stability analysis in~\cite{yin2020stability} provides stability certificates and estimates the region of attraction (when global stability tests are infeasible).

In accordance with~\cite{yin2020stability}, the followings are introduced to obtain the Lyapunov condition for the neural network controller. 
The state $x_*$ is an equilibrium point of the feedback system with input $u_*$ if the following conditions hold:
\begin{subequations}
\label{Eq:DefEquilibriumPoint}
\begin{align}
x_*&=A_Gx_*+B_Gu_*,\\
u_*&=\pi(x_*).
\end{align}
\end{subequations}

Suppose that $(x_*, u_*)$ satisfies Eq.~(\ref{Eq:DefEquilibriumPoint}). 
Then, the equilibrium point $x_*$ can be propagated throu the neural network to obtain equilibrium values $v^i_*$, $w^i_*$ for the inputs/outputs of each activation function $(i=1,...,\ell)$, and thus, $(x_*,u_*,v_*,w_*)$ is an equilibrium point of Eqs.~(\ref{Eq:Plant}) and (\ref{Eq:OrdinalNeuralController}) if:

\begin{subequations}
\label{Eq:EquilibriumPoint}
\begin{align}
x_*&=A_Gx_*+B_Gu_*, \\
\left[\hspace{-1mm}
\begin{array}{cc}
u_*\\
v_*
\end{array}
\hspace{-1mm}\right]
&=N
\left[\hspace{-1mm}
\begin{array}{cc}
x_*\\
w_*\\
1
\end{array}
\hspace{-1mm}\right],\\
w_*&=\phi(v_*).
\end{align}
\end{subequations}
where $\phi$ is the combined nonlinearity by stacking the activation functions. 
The matrix $N$ is defined by
\begin{eqnarray}
N:=\left[
\begin{array}{c|cccc|c}
0&0&0&\ldots&W^{\ell+1}&b^{\ell+1}\\\hline
W^1&0&\ldots&0&0&b^1\\
0&W^2&\ldots&0&0&b^2\\
\vdots&\vdots&\ddots&\vdots&\vdots&\vdots\\
0&0&\ldots&W^\ell&0&b^\ell
\end{array}
\right]
:=
\left[
\begin{array}{ccccc}
N_{ux}&N_{uw}&N_{ub}\\
N_{vx}&N_{vw}&N_{vb}
\end{array}
\right].
\end{eqnarray}

Finally, the ellipsoied is defined for $P\in\mathbb{S}^n_{++}$, $x_*\in\mathbb{R}^n$:
\begin{eqnarray}
\label{Eq:Ellipsoid}
\mathcal{E}(P,x_*):=\{x\in\mathbb{R}^n: (x-x_*)^TP(x-x_*)\le1 \}.
\end{eqnarray}

With the above definitions, the following property and the theorem is obtained by Ref.~\cite{yin2020stability}.
\begin{property}{\bf (Local Sector~\cite{yin2020stability})}{\it\ 
Let $v_*\in\mathbb{R}^{n_\phi}$ be an equilibrium value of the activation input and $v^1_*\in\mathbb{R}^{n_1}$ be the corresponding value at the first layer.
Select $\underline{v}^1, \overline{v}^1\in\mathbb{R}^{n_1}$ with $v^1_*\in[\underline{v}^1, \overline{v}^1]$.
Let $\hat{v}_\phi(v^1)\in\mathbb{R}^{n_\phi}$ denote the activation input generated by Eq.~(\ref{Eq:OrdinalNeuralControllerb}) from the input $v^1\in\mathbb{R}^{n_1}$ at the first layer.
There exist $\alpha_\pi, \beta_\phi\in\mathbb{R}^n_\phi$ such that $\phi$ satisfies the offset local sector around the point $(v_*, \phi(v_*))$ for all $\hat{v}_\phi(v^1)$ with $v^1\in[\underline{v}^1, \overline{v}^1]$.
}
\end{property}

\begin{theorem}
\label{Th:LyapunovCondition}
{\bf (Lyapunov Condition~\cite{yin2020stability})}
Consider the feedback system of plant $G$ in~(\ref{Eq:Plant}) and the neural network controller $\pi$ in~(\ref{Eq:OrdinalNeuralController}) with equilibrium point $(x_*, u_*, v_*, w_*)$ satisfying~(\ref{Eq:EquilibriumPoint}).
Let $\overline{v}^1\in\mathbb{R}^{n_1}$, $\underline{v}^1:=2v^1_*-\overline{v}^1$, and $\alpha_\phi, \beta\in\mathbb{R}^{n_\phi}$ be given vectors satisfying Property 1 for the neural network. Denote the $i^{th}$ row of the first weight $W^1$ by $W^1_i$ and define matrices
\begin{eqnarray}
R_V:=
\left[\hspace{-1mm}
\begin{array}{cc}
I_{n_G}&0_{n_G\times n_\phi}\\
N_{ux}&N_{uw}
\end{array}
\hspace{-1mm}\right],\quad and \quad
R_\phi:=
\left[\hspace{-1mm}
\begin{array}{cc}
N_{vx}&N_{vw}\\
0_{n_\phi\times n_G}&I_{n_\phi}
\end{array}
\hspace{-1mm}\right].
\end{eqnarray}
Define further matrices
\begin{eqnarray}
\Psi_\phi:= 
\left[
\begin{array}{ccccc}
diag(\beta_\phi)&-I_{n_\phi}\\
-diag(\alpha_\phi)&I_{n_\phi}
\end{array}
\right], \quad and \quad
M_\phi(\lambda):= 
\left[
\begin{array}{ccccc}
0_{n_\phi}&diag(\lambda)\\
diag(\lambda)&0_{n_\phi}
\end{array}
\right],
\end{eqnarray}
where $\lambda\in\mathbb{R}^{n_\phi}$ with $\lambda\ge0$. 
If there exists a matrix $P\in\mathbb{S}^{n_G}_{++}$ and vector $\lambda\in\mathbb{R}^{n_\phi}$ with $\lambda\ge0$ such that
\begin{eqnarray}
\label{Eq:TheLMI-1}
R_V^T
\left[
\begin{array}{ccccc}
A_G^TPA_G-P&A_G^TPB_G\\
B_G^TPA_G&B_G^TPB_G
\end{array}
\right]
R_V
+R_\phi^T\Psi_\phi^TM_\phi(\lambda)\Psi_\phi R_\phi<0, \\
\left[
\begin{array}{ccccc}
(\overline{v}_i^1-v_{*,i}^1)^2&W_i^1\\\label{Eq:TheLMI-2}
W_i^{1T}&P
\end{array}
\right]
\geq 0, \quad i=1, ..., n_1,
\end{eqnarray}
then: (i) the feedback system consisting of $G$ and $\pi$ is locally stable around $x_*$, and (ii) the set $\mathcal{E}(P, x_*)$, defined by Eq.~(\ref{Eq:Ellipsoid}) is an inner-approximation to the ROA.
\end{theorem}

\subsection{Spectral Normalization}
The spectral normalization~\cite{miyato2018spectral} has been proposed for the generative adversarial networks (GANs)~\cite{goodfellow2014generative}.
The training of GANs are stabilized by regularizing the Lipschitz norm of the network layers.
The spectral norm of the matrix $W$ is defined by
\begin{equation}
\sigma(W) = \sup_{h \neq 0} \frac{||Wh||_2}{||h||_2} = \max_{||h||_2 \le 1} ||Wh||_2,
\end{equation}
which is equivalent to the largest singular value of $W$.
The spectral normalization is defined by
\begin{equation}
\bar W_{SN} = W / \sigma(W).
\end{equation}

\section{Method}
\label{Sec:Method}
In this section, the relationship between the spectral normalization and the stability of the feedback system is described.
The original spectral normalization~\cite{miyato2018spectral} simply normalizes the weight matrix such that $\bar W_{SN} = W / \sigma(W)$, which means that the spectral norm of the normalized matrix is given by $\sigma(\bar{W}_{SN})=1$. 
In this article, however, the following weight normalization is introduced for the flexibility of the policy design: 
\begin{eqnarray}
\label{Eq:SNwithDelta}
\hat{W} := \cfrac{\delta}{\sigma(W)}W, \quad \delta>0,
\end{eqnarray}
where $\hat{W}$ and $\delta$ are respectively the normalized weight matrix and the positive constant.
By introducing the parameter $\delta$, it allows us to arbitrarily normalize the spectral norm of the weight matrix with $\sigma(\hat{W})= \delta$.
The key point is to appropriately select the parameter $\delta$ for ensuring the stability of the feedback system. 
In Section~\ref{SubSec:Stability}, the method for ensuring the global stability of the feedback system is described. 
In Section~\ref{SubSec:LocalStability}, the more practical method is proposed for improving control performance while ensuring the local stability of the feedback system.

In the following, the bias term of each layer is omitted for simplicity (i.e. the equilibrium points are $x^*=0$ and $u^*=0$). 
In accordance with~\cite{yin2020stability}, the nonlinear operation of the activation function is isolated from the linear operation.
Define $v^i$ as the input to the activation function $\phi^i$ as follows:
\begin{eqnarray}
v^i(k):=W^iw^{i-1}(k), \quad i=1, ..., \ell.
\end{eqnarray}

The neural network controller is thus represented as follows:
\begin{subequations}
\label{Eq:NoBiasNeuralController}
\begin{align}\label{Eq:NoBiasNeuralControllera}
w^0(k) &= x(k); \\\label{Eq:NoBiasNeuralControllerb}
v^i(k) &= W^i w^{i-1}(k), \quad i = 1, ..., \ell; \\\label{Eq:NoBiasNeuralControllerc}
w^i(k) &= \phi^i (v^i(k)), \quad i = 1, ..., \ell;  \\\label{Eq:NoBiasNeuralControllerd}
u(k) &= W^{\ell+1} w^\ell(k).
\end{align}
\end{subequations}

\subsection{Pre-Guaranteed RL}
\label{SubSec:Stability}
The method described in this section is a standard approach: ensuring the $\mathcal{L}_2$ gain of the feedback system is less than 1 by normalizing all weight matrices in the neural network. 
This method is called {\it pre-guaranteed RL} in this article.
Although the {\it pre-guaranteed RL} is essentially the same as the original spectral normalization~\cite{miyato2018spectral} except for introducing the parameter $\delta$ as defined in Eq.~(\ref{Eq:SNwithDelta}), the importance of this introduction is explained as follows.

Let us consider the linear operation in the hidden layers (i.e. Eq.~(\ref{Eq:NoBiasNeuralControllerb})).
The spectral normalization defined by Eq.~(\ref{Eq:SNwithDelta}) is performed for each weight matrix of the hidden layers: 
\begin{equation}
\label{Eq:WeightNormalization}
\hat{W}^i = \cfrac{\delta^i}{\sigma(W^i)}W^i, \quad \delta^i>0, \quad i=1, ..., \ell,
\end{equation}
where $\hat{W}^i$ and $\delta^i$ are respectively the normalized weight matrix and the positive constant for the $i^{th}$ layer.
This means that with the tuning parameter $\delta^i$, the spectral norm of the weight matrix is given by $\sigma(\hat{W}^i)= \delta^i$. 
Subsequently, let us define the diagonal matrix $D^i_x$, whose diagonal elements are the gain of the activation function for a given vector $x$, and define the matrix $W_x:=D^{\ell}_x \hat{W}^\ell D^{\ell-1}_x \hat{W}^{\ell-1}\cdots D^1_x\hat{W}^1$ corresponding to the operation of Eqs.~(\ref{Eq:NoBiasNeuralControllera})-(\ref{Eq:NoBiasNeuralControllerc}), i.e., $W_x: x\to w^\ell$.
Note that the spectral norm of the diagonal matrix is $\sigma(D^i_x)\leq 1$\footnote[2]{This property is satisfied for the activation functions which are commonly used in the neural networks, e.g. tanh, ReLU, and sigmoid. See Ref.~\cite{kretchmar2000synthesis} for the details in the case of tanh, which helps one understand this article.}, and the following inequality is obtained:
\begin{eqnarray}
\label{Eq:SNforHiddenLayer}
\sigma(W_x)\le \sigma(D^\ell_x)\cdot \sigma(\hat{W}^\ell)\cdot \sigma(D^{\ell-1}_x) \cdot \sigma(\hat{W}^{\ell-1})\cdots \sigma(D^1_x)\cdot \sigma(\hat{W}^1)\le \prod_{i=1}^\ell \delta^i.
\end{eqnarray}
Next, the spectral normalization is performed for the linear operation of Eq.~(\ref{Eq:NoBiasNeuralControllerd}), i.e.
\begin{eqnarray}
\hat{W}^{\ell+1}=\frac{\delta^{\ell+1}}{\sigma(W^{\ell+1})}W^{\ell+1}, \quad \delta^{\ell+1}>0.
\end{eqnarray}
The neural network controller represented in  Eq.~(\ref{Eq:NoBiasNeuralController}) is then re-formulated by
\begin{subnumcases}
  {\pi_{pre}:\quad}
\ w^0(k)&$=\ x(k)$; \\
\ v^i(k)&$=\ \hat{W}^i w^{i-1}(k), \quad i = 1, ..., \ell;$ \\
\ w^i(k) &$=\ \phi^i(v^i(k)), \quad i = 1, ..., \ell$; \\
\ u(k) &$=\ \hat{W}^{\ell+1} w^\ell(k)$,
\end{subnumcases}
and the spectral norm of the matrix $W^{\ell+1}W_x$, which corresponds to the mapping $\pi_{pre}: x\to u$, is given by 
\begin{eqnarray}
\sigma(W^{\ell+1}W_x)\le \sigma(W^{\ell+1})\cdot \sigma(W_x) \le \prod_{i=1}^{\ell+1} \delta^i.
\end{eqnarray}
This means that the output of the neural network controller (and also the input/output of the activation functions in the hidden layers) is bounded as follows:
\begin{eqnarray}
\label{Eq:GainNonlinear}
||u(k)||_2\le\delta^{\ell+1}||w^\ell(k)||_2 \le \delta^{\ell+1}||v^\ell(k)||_2\le \delta^{\ell+1}\delta^\ell||w^{\ell-1}(k)||_2\le \cdots \le ||x(k)||_2\prod_{i=1}^{\ell+1} \delta^i.
\end{eqnarray}
To investigate the stability of the feedback system, it is sufficient to have the input-output relation of the neural network controller, i.e.
\begin{eqnarray}
\label{Eq:NNOutput}
||u(k)||_2\le \overline{\sigma}_\pi||x(k)||_2,
\end{eqnarray}
where
\begin{eqnarray}
\label{Eq:GammaPi}
\overline{\sigma}_\pi&:=& \prod_{i=1}^{\ell+1} \delta_i.
\end{eqnarray}

Finally, consider the $\mathcal{L}_2$ gain of the neural network controller:
\begin{eqnarray}\nonumber
\label{Eq:L2gainNN}
\gamma_\pi &=& \sup_{x\neq0}\cfrac{||u||_{\mathcal{L}_2}}{||x||_{\mathcal{L}_2}} \nonumber\\
&=&\sup_{x\neq0} \cfrac{ \sqrt{\sum_{k=0}^\infty ||u(k)||^2_2}}{\sqrt{\sum_{k=0}^\infty||x(k)||^2_2}}\nonumber\\
&\le& \sup_{x\neq0} \cfrac{ \sqrt{\sum_{k=0}^\infty {\overline{\sigma}_\pi}^2||x(k)||^2_2}}{\sqrt{\sum_{k=0}^\infty||x(k)||^2_2}}\nonumber\\
&=&\sup_{x\neq0} \cfrac{\overline{\sigma}_\pi \sqrt{\sum_{k=0}^\infty ||x(k)||^2_2}}{\sqrt{\sum_{k=0}^\infty||x(k)||^2_2}}\nonumber\\
&=&\overline{\sigma}_\pi.
\end{eqnarray}
This means that the mapping $\pi_{pre}:x\to u$ has a finite $\mathcal{L}_2$ gain and it is less than or equal to $\overline{\sigma}_\pi$. 
The upper bound of the $\mathcal{L}_2$ gain of the neural network controller can be calculated by Eq.(\ref{Eq:GammaPi}). 
Thus, the stability condition of the feedback system as shown in Fig.~\ref{Fig:FeedbackNN} is given as follows.

\begin{theorem}
\label{Th:SNStabilityCondition}
{\bf (Stability Condition)}
Consider the feedback system as shown in Fig.~\ref{Fig:FeedbackNN}. 
Suppose that the plant $G$ has a finite $\mathcal{L}_2$ gain and it is less than or equal to $\gamma_G$. 
Suppose further that each weight matrix of the neural network controller is normalized as in Eq.~(\ref{Eq:SNwithDelta}). 
Assuming the feedback connection is well defined, the feedback system is finite-gain $\mathcal{L}_2$ stable by the small-gain theorem if
\begin{eqnarray}
\label{Eq:StabilityCondition}
\overline{\sigma}_\pi\gamma_G<1,
\end{eqnarray}
where $\overline{\sigma}_\pi$ is defined by Eq.~(\ref{Eq:GammaPi}).
\end{theorem}

\begin{proof}
Let $\gamma_\pi$ be the $\mathcal{L}_2$ gain of the neural notwork controller.
From Eq.~(\ref{Eq:L2gainNN}), we have the upper bound with $\gamma_\pi \le \overline{\sigma}_\pi$.
The $\mathcal{L}_2$ gain of the feedback system can be bounded by $\gamma_\pi\gamma_G \le \overline{\sigma}_\pi \gamma_G$.
Therefore, $\overline{\sigma}_\pi \gamma_G \le 1 \Rightarrow \gamma_\pi \gamma_G < 1$, and then, the feedback system is finite-gain $\mathcal{L}_2$ stable by small-gain theorem.
\end{proof}

To summarize the {\it pre-guaranteed RL}, the spectral normalization as in Eq.~(\ref{Eq:SNwithDelta}) is proposed and the stability condition of the feedback system is derived based on this spectral normalization. 
The stability is ensured by normalizing the weight matrices of all layers under the condition of $\overline{\sigma}_\pi<1/\gamma_G$.
This method can be applied to a lot of existing RL algorithms since it does not require any change of the basic framework of those algorithms. 
However, there exist the following limitations:
\begin{enumerate}
\renewcommand{\labelenumi}{(\alph{enumi})}
\item only applicable for a plant $G$ which has a finite $\mathcal{L}_2$ gain, 
\item strict limitation for the neural networks due to Eq.~(\ref{Eq:StabilityCondition}).
\end{enumerate}
Regarding (a), there exist cases that the {\it pre-guaranteed RL} cannot be used for traditional RL problems, e.g., the linearized model of the inverted pendulum at the equilibrium point $(\theta^*, q^*)=(0, 0)$, where $\theta$ and $q$ are respectively the angle and the angular velocity, does not have a finite $\mathcal{L}_2$ gain.
Regarding (b), the designed policy may not satisfy the performance requirement due to the constraint of the stability condition. 
As the gain of the plant become larger, the smaller gain of the policy is required to achieve the stability condition, resulting in an insufficient performance of the designed policy. 
To overcome these difficulties, in the next section, the method shall be proposed. 

\subsection{Post-Guaranteed RL}
\label{SubSec:LocalStability}
Although the {\it pre-guaranteed RL} explicitly includes the stability condition in the training of the policy by using the spectral normalization, the stability condition in Theorem~\ref{Th:SNStabilityCondition} may impose severe constraints for the policy. 
Moreover, it is easily confirmed that the {\it pre-guaranteed RL} cannot be used for the plant which has a large $\mathcal{L}_2$ gain since the $\mathcal{L}_2$ gain of the neural network controller must be set to nearly~0. 

If one intends to guarantee a global stability, it often leads to the poor performance of the feedback system. 
In contrast, if one pays attention to a local stability, it may improve the control performance of the feedback system. 
In order to investigate the local stability, let us consider the region of attraction (ROA) of the feedback system defined by
\begin{eqnarray}
\mathcal{R}:=\{x_0 \in \mathbb{R}^{n_G}: \lim_{k \to \infty} \chi(k; x_0) = x_* \},
\end{eqnarray}
where $\chi(k; x_0)$ is the solution to the feedback system at time $k$ from the initial condition $x(0)=x_0$. 
The ROA is the set such that all the initial conditions converge to the equilibrium point $x_*$ as $k\to\infty$~\cite{khalil2002nonlinear}.
The goal is to obtain the policy which has a larger ROA than a design requirement.
In this article, the method to ensure the stability in the sense of the ROA is called {\it post-guaranteed RL} as explained below.

In order to avoid the limitations due to the stability condition of Eq.~(\ref{Eq:StabilityCondition}), in the {\it post-guaranteed RL}, the spectral normalization shall be applied only for each hidden layer but for the output layer.
The neural network controller in the {\it post-guaranteed RL} is then represented as follows:
\begin{subnumcases}
  {\pi_{post}:\quad}
\ w^0(k)&$=\ x(k)$; \\
\ v^i(k)&$=\ \hat{W}^i w^{i-1}(k), \quad i = 1, ..., \ell$; \\
\ w^i(k) &$=\ \phi^i(v^i(k)), \quad i = 1, ..., \ell$; \\
\ u(k) &$=\ W^{\ell+1} w^\ell(k)$.
\end{subnumcases}
Note that the stability of the feedback system is not ensured by normalizing only each hidden layer.
However, this is a compromise such that the policy can be obtained with improved performance (and the optimization algorithm can be applied even for the system which does not have a finite $\mathcal{L}_2$ gain). 
After obtaining the policy, an a-posteriori analysis shall be performed to confirm the local stability with its ROA (i.e., the stability of the feedback system is a-posteriorly ensured).
Specifically, the {\it post-guaranteed RL} may improve the feasibility of the a-posteriori stability tests and enlarge the ROA. 

\begin{figure}[!!b]
\begin{center}
\includegraphics[width=8cm]{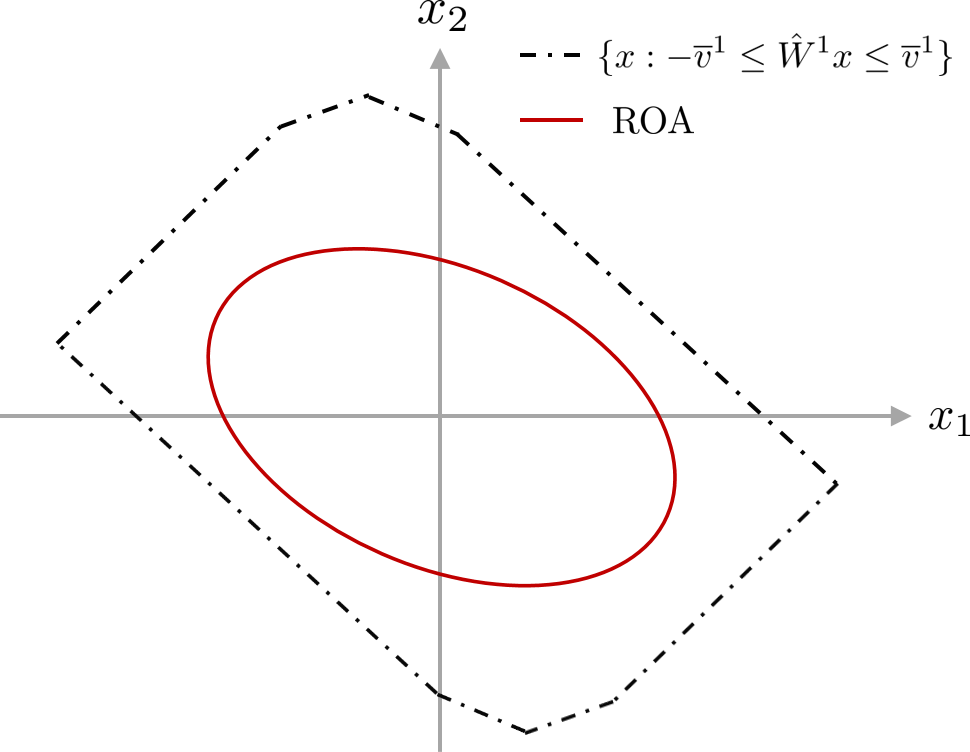}
\caption{Example of the ROA for the feedback system with the state $x=[x_1, x_2]^T$.}
\label{Fig:ROA_image}
\end{center}
\vspace{-5mm}
\end{figure}

In accordance with~\cite{yin2020stability}, let $v^1_*\in\mathbb{R}^{n_1}$ be the equilibrium point at the first layer. 
Assume the bounds $\underline{v}^1$ and $\overline{v}^1$ with $\underline{v}^1\le v^1_* \le \overline{v}^1$ and $\underline{v}^1=2v^1_*-\overline{v}^1$.
Note that as explained in~\cite{yin2020stability}, the assumed bounds $[\underline{v}^1, \overline{v}^1]$ can be used to obtain the local sector bounds $[\alpha^i, \beta^i]$ for all the nonlinear activation functions $\phi^i$ ($i=1,..., \ell$) in the neural network.
In this article, the bias in the neural network is set to zero, and thus, $v_*=0$ and $\overline{v}^1=-\underline{v}^1$.
Therefore, if the stability condition formulated by the LMIs in Theorem~\ref{Th:LyapunovCondition} is feasible, the ROA exists in the following region:
\begin{eqnarray}
\label{Eq:ROAwithSN}
\mathcal{E}(P,x_*=0)\subseteq\{x: -\overline{v}^1\le \hat{W}^1x \le \overline{v}^1 \}.
\end{eqnarray}

For understanding the relationship between the set of the ellipsoid $\mathcal{E}(P, x_*=0)$ and the spectral normalization defined by Eq.~(\ref{Eq:SNwithDelta}), let us consider a simple example of the plant with the state $x=[x_1, x_2]^T\in \mathbb{R}^2$. 
In this example, it is assumed that $v^1\in[\underline{v}^1, \overline{v}^1]$ with $\overline{v}^1=-\underline{v}^1=\mu^1\times\mathbf{1}_{2\times 1}$, where $\mu^1$ is the positive scalar value and $\mathbf{1}_{2\times1}$ is the all-ones vector.
Figure~\ref{Fig:ROA_image} shows an example of the regions given by Eq.~(\ref{Eq:ROAwithSN}), which means that the set of the ellipsoid $\mathcal{E}(P, x_*)$ exists in the region of $\{x: -\overline{v}^1\le \hat{W}^1x \le \overline{v}^1\}$. 
In other words, the region of $\{x: -\overline{v}^1\le \hat{W}^1x \le \overline{v}^1\}$ that the ROA exists is determined by two parameters $\delta^1$ and $\mu^1$.

\begin{figure}[!!t]
\begin{center}
\includegraphics[width=15cm]{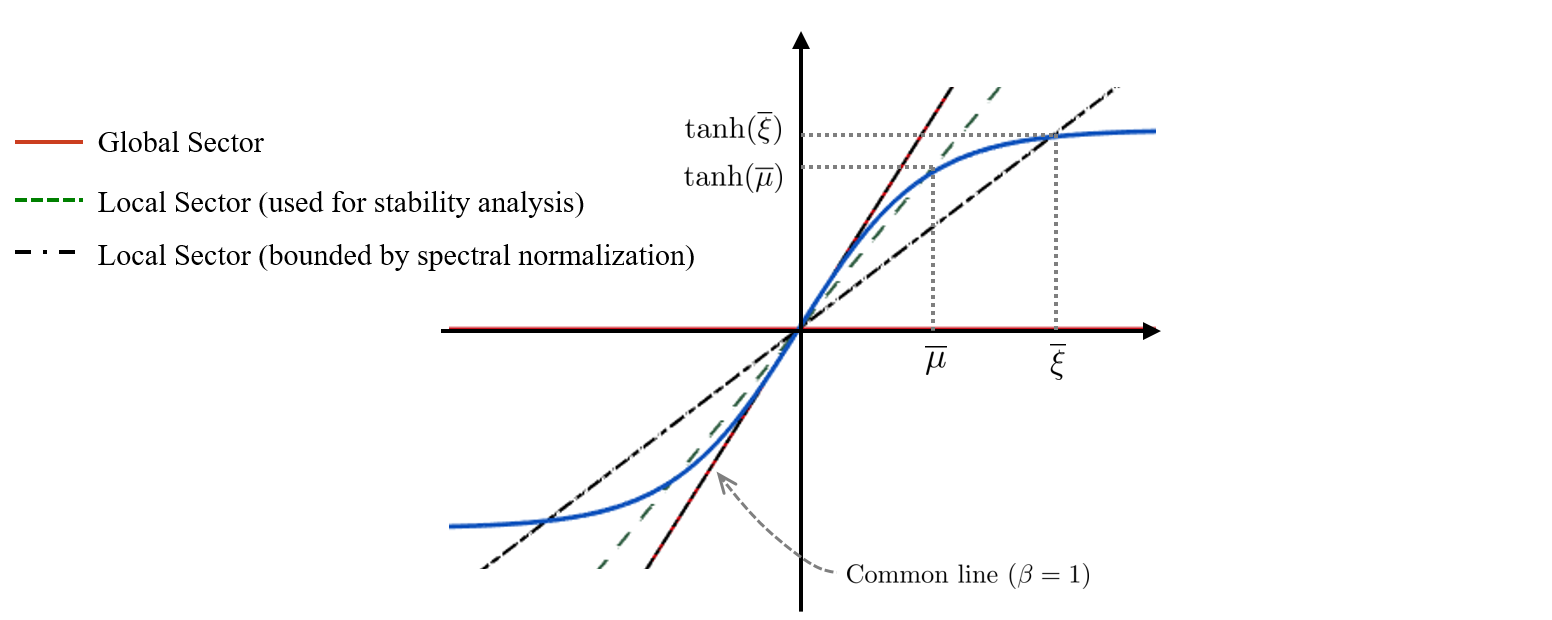}
\caption{Three types of regions of sector bounded nonlinearity.}
\label{Fig:LocalSector}
\end{center}
\vspace{-5mm}
\end{figure}

In the stability analysis, we need to set $\mu^1$ to a smaller value (i.e. assume tighter local sectors) to make the LMIs in Theorem~\ref{Th:LyapunovCondition} feasible. 
This means that the region of $\{x: -\overline{v}^1\le \hat{W}^1x \le \overline{v}^1\}$ cannot be enlarged with $\mu^1$ due to the feasibility constraint. 
In the {\it post-guaranteed RL}, the effectiveness of introducing the parameters $\delta^i$, $i=1,...,\ell$, is explained as follows. 
First, setting $\delta^1$ to a small value enlarges the region of $\{x: -\overline{v}^1\le \hat{W}^1x \le \overline{v}^1\}$ under a small value of $\mu^1$, which provides a potential to obtain a larger ROA. 
Second, from Eq.~(\ref{Eq:SNforHiddenLayer}), the input to the nonlinear activation function is limited. Thus, the assumed sectors for all nonlinear functions can be smaller, resulting in the improvement of the feasibility in the stability analysis formulated by the LMIs\footnote{The sectors after the second layer can be obtained with a forward propagation from the one of the first layer, see~\cite{yin2020stability} for the details. 
If the weights of the $i^{th}$ hidden layers are unbounded, the calculated sectors may tend to be large, resulting in the infeasibility of the stability condition formulated by the LMIs in Theorem~\ref{Th:LyapunovCondition}.}. 
Note that the feasibility analysis of LMIs has been well investigated in the field of control engineering, see Ref.~\cite{shimomura2005gain} for the details of the relationship between the region size and the feasibility of LMIs in conjunction with pre-guaranteed and post-guaranteed approach. 
Figure~\ref{Fig:LocalSector} shows three types of regions of sector bounded nonlinearity, in which $\tanh$ is selected as the nonlinear activation function. 
The red solid line shows the global sector defined by $[\alpha, \beta]$ with $\alpha=0$ and $\beta=1$, the green dashed line the local sector with $\alpha=\tanh(\overline{\mu})/\overline{\mu}$ and $\beta=1$ used for the stability analysis, and the black dash-dot line the local sector with $\alpha=\tanh(\overline{\xi})/\overline{\xi}$ and $\beta=1$ bounded by the spectral normalization. 
From Fig.~\ref{Fig:LocalSector}, it can be confirmed that the spectral normalization enforces the smaller sector bound. 
On the other hand, too smaller values of $\delta^i$, $i=1,...,\ell$, strictly limit the nonlinearities of the activation functions, which means that too smaller values of $\delta^i$ provide the poor performance of the policy.

To summarize the {\it post-guaranteed RL}, we can make sectors arbitrary small for making the LMIs feasible while enlarging the ROA by properly choosing parameters $\overline{v}^1$ and $\delta^i$, $i=1,...,\ell$.
In other words, the potential size of the ROA can be changed (for satisfying a design requirement) with trade-off between stability and performance for the neural network controller.

\section{Experiments}
\label{Sec:Experiments}
Two numerical experiments are performed for discrete-time LTI systems: one is the inverted pendulum and the other is the longitudinal motion of aircraft. 
The details of the environments used in this article are provided in Appendix~\ref{Sec:EnvDetails}. 
The policy is modeled by a fully-connected multi-layer perceptron with $\rm tanh$ as the activation function, which is trained through the policy optimization algorithm using PPO~\cite{schulman2017proximal}. 
In this article, PPO is selected as the baseline RL algorithm. 
For investigating the effectiveness of the spectral normalization, the {\it pre-guaranteed RL} and the {\it post-guaranteed RL} are compared with the baseline PPO (but the {\it pre-guaranteed RL} is tested only for the stable system, i.e. the aircraft control task).

\subsection{Inverted Pendulum}
The inverted pendulum is the traditional benchmark problem for RL algorithms. 
The state is $x=[\theta, \dot{\theta}]$, where $\theta$ and $\dot{\theta}$ are respectively the angle (rad) and the angular velocity (rad/s). 
The input is $u=\tau$, where $\tau$ is the torque ($\rm N\cdot m$). 
For the experiment, the nonlinear model is linearized around the equilibrium point $x^*=[0, 0]^T$ in accordance with~\cite{okawa2019control}. 
Note that the linearized model around the equilibrium point is unstable and does not have a finite $\mathcal{L}_2$ gain, which means that the {\it pre-guaranteed RL} cannot be used for this problem. 
The policy network has two hidden layers of 64 units.

\begin{figure}[!!t]
  \begin{center}
    \begin{tabular}{c}

      \begin{minipage}{0.5\hsize}
        \begin{center}
          \includegraphics[clip, width=8.3cm]{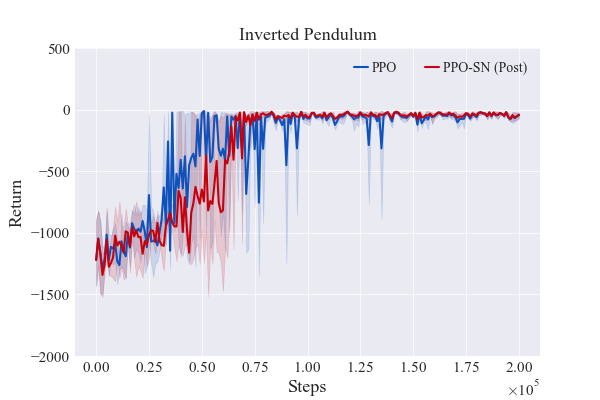}
          \hspace{1.6cm} (a) Learning curves.
        \end{center}
      \end{minipage}

      \begin{minipage}{0.5\hsize}
        \begin{center}
          \includegraphics[clip, width=8.3cm]{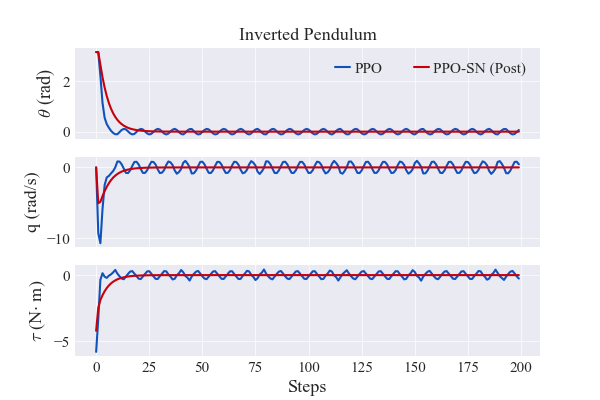}
          \hspace{1.6cm} (b) Trajectories.
        \end{center}
      \end{minipage} \\

      \begin{minipage}{0.5\hsize}
        \begin{center}
          \includegraphics[clip, width=8.2cm]{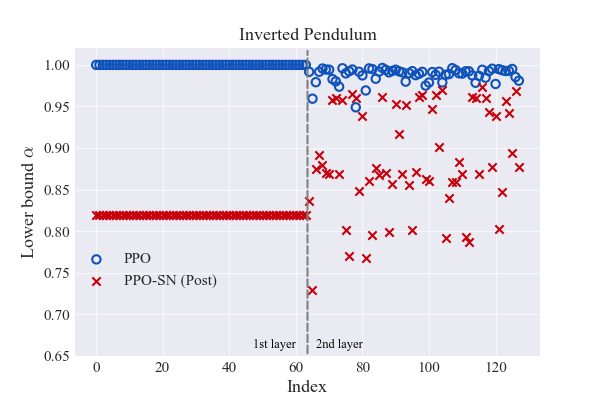}
          \hspace{2.6cm} (c) Calculated lower bounds.
        \end{center}
      \end{minipage}
      
      \begin{minipage}{0.5\hsize}
        \begin{center}
          \includegraphics[clip, width=8.2cm]{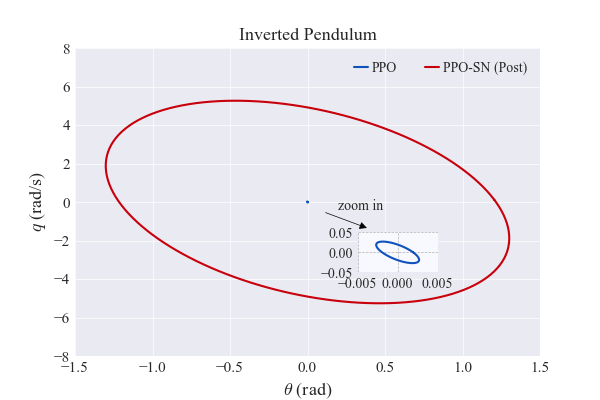}
          \hspace{2.6cm} (d) ROA.
        \end{center}
      \end{minipage}

    \end{tabular}
    \caption{Comparison between PPO and PPO-SN (Post) on the inverted pendulum task.}
    \label{Fig:ResultPendulum}
  \end{center}
\end{figure}

Figure~\ref{Fig:ResultPendulum} shows the result of the inverted pendulum task, in which the blue line/marker shows the baseline PPO and the red line/marker shows the PPO via the spectral normalization (SN) using the {\it post-guaranteed RL}. 
Figure~\ref{Fig:ResultPendulum}-(a) shows the learning curves, where the solid line corresponds to the average and the shaded region to the minimum/maximum returns of evaluation rollouts without exploration noise over the three random seeds. 
Figure~\ref{Fig:ResultPendulum}-(b) shows the trajectories obtained by the policy without exploration noise, where the initial states of the system are set to $x_0=[\pi, 0]^T$. 
Figure~\ref{Fig:ResultPendulum}-(c) shows the calculated lower bounds for the a-posteriori stability analysis, and Fig.~\ref{Fig:ResultPendulum}-(d) shows the obtained ROA. 
The results shown in Figs.~\ref{Fig:ResultPendulum} (b) to (d) are the case of the first random seed in the experiments.

From Figs.~\ref{Fig:ResultPendulum}-(a) and (b), PPO-SN (Post) performs comparably to the baseline PPO and achieves the control task with more smooth trajectories. 
Remarkable difference between PPO and PPO-SN (Post) is seen in the results of the a-posteriori stability analysis, see Figs.~\ref{Fig:ResultPendulum}-(c) and (d). 
In the case of PPO, the small sector bounds (i.e. $\alpha$ is set to nearly $\beta=1$) for the first layer are assumed in order to make the LMIs feasible, which corresponds to shrink the size of the ROA. 
On the other hand, in the case of PPO-SN (Post), the larger sector bounds than those of PPO can be assumed while making the LMIs feasible since the spectral norm of the weight matrices in the first/second layers are bounded (see also Table~\ref{tab:WeightComparisonPendulum}). Thus, by the spectral normalization, the larger ROA can be obtained in the a-posteriori stability analysis.

\begin{figure}[!!t]
  \begin{center}
    \begin{tabular}{c}

      \begin{minipage}{0.5\hsize}
        \begin{center}
          \includegraphics[clip, width=8.3cm]{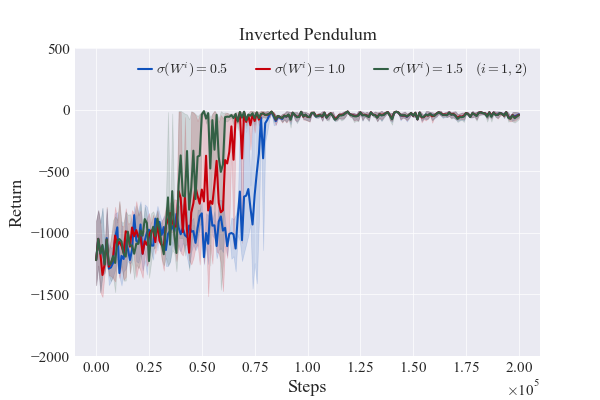}
          \hspace{1.6cm} (a) Learning curves.
        \end{center}
      \end{minipage}

      \begin{minipage}{0.5\hsize}
        \begin{center}
          \includegraphics[clip, width=8.3cm]{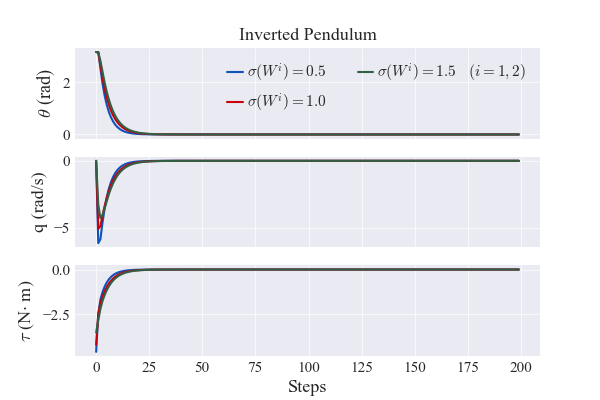}
          \hspace{1.6cm} (b) Trajectories.
        \end{center}
      \end{minipage} \\

      \begin{minipage}{0.5\hsize}
        \begin{center}
          \includegraphics[clip, width=8.2cm]{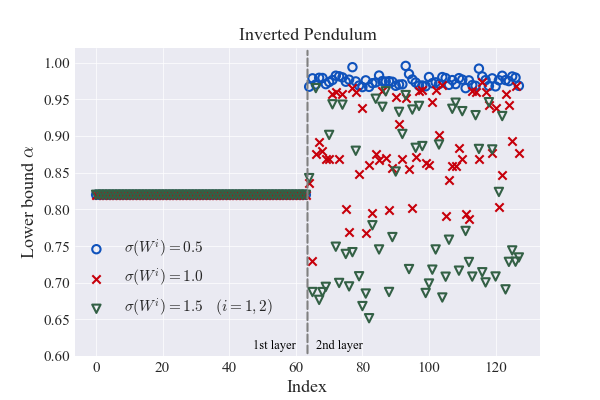}
          \hspace{2.6cm} (c) Calculated lower bounds.
        \end{center}
      \end{minipage}
      
      \begin{minipage}{0.5\hsize}
        \begin{center}
          \includegraphics[clip, width=8.2cm]{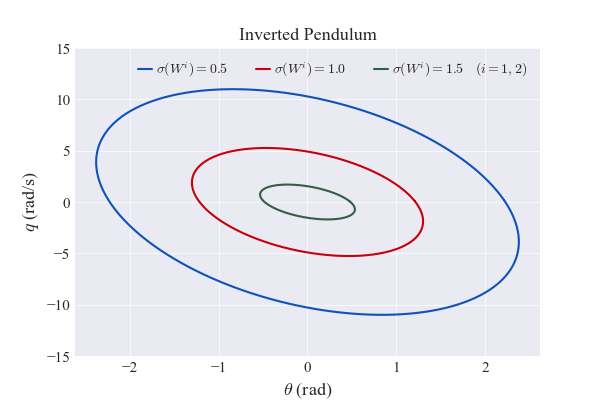}
          \hspace{2.6cm} (d) ROA.
        \end{center}
      \end{minipage}

    \end{tabular}
    \caption{Comparison of PPO-SN (Post) by norm size on the inverted pendulum task.}
    \label{Fig:ResultPendulum_comparison_by_norm}
  \end{center}
\end{figure}

Figure~\ref{Fig:ResultPendulum_comparison_by_norm} shows the result of the comparison by the size of the spectral norm (i.e. the parameters $\delta^i$, $i=1,2$) in the case of PPO-SN (Post). 
The blue line/marker shows the case of $\delta^i=0.5$, the red $\delta^i=1.0$, and the green $\delta^i=1.5\ (i=1,2)$. 
Although more sample efficiency is seen by increasing the scale size of the spectral norm (Fig.~\ref{Fig:ResultPendulum_comparison_by_norm}-(a)), the trajectories obtained after the training on the total step $2\times10^5$ are almost similar (Fig.~\ref{Fig:ResultPendulum_comparison_by_norm}-(b)). 
From Figs.~\ref{Fig:ResultPendulum_comparison_by_norm}-(c) and (d), the larger ROA can be obtained by setting $\delta^i$, $i=1, 2$, to the smaller value. 
Note that the sector sizes for the first layer are set to the same in the stability analysis of each case. 
From these results, it is confirmed that the trade-off between performance and stability due to the spectral normalization. 
Table~\ref{tab:WeightComparisonPendulum} shows the spectral norm of the weight matrices obtained on the inverted pendulum task. 
Regarding the weight matrices of the first/second layers, the norm size of PPO becomes larger than that of PPO-SN (Post). 
The difference in the norm size leads to the difference in the calculated lower bounds of the second layer in the a-posteriori stability analysis as explained above. 
From these results, the spectral normalization improves the feasibility of the a-posteriori analysis and enlarges the ROA.

\begin{table}[!!t]
\begin{center}
\caption{Spectral norm of the weight matrices obtained on the inverted pendulum task. The spectral normalization is performed for hidden layers in PPO-SN (Post).}
\vspace{2mm}
\begin{tabular}{cccccc}\hline
& PPO &\multicolumn{3}{c}{PPO-SN (Post)}\\
&-&$\delta=0.5$ & $\delta=1.0$ & $\delta=1.5$ \\\hline
$W_1$ & 3.423 & 0.5000 & 1.000 & 1.500 \\
$W_2$ & 4.201 & 0.5000 & 1.000 & 1.500 \\
$W_3$ & 1.171 & 6.836  & 1.882 & 1.123 \\ \hline    
\end{tabular}
\label{tab:WeightComparisonPendulum}
\end{center}
\end{table}

\begin{figure}[!!t]
  \begin{center}
    \begin{tabular}{c}

      \begin{minipage}{0.5\hsize}
        \begin{center}
          \includegraphics[clip, width=8.3cm]{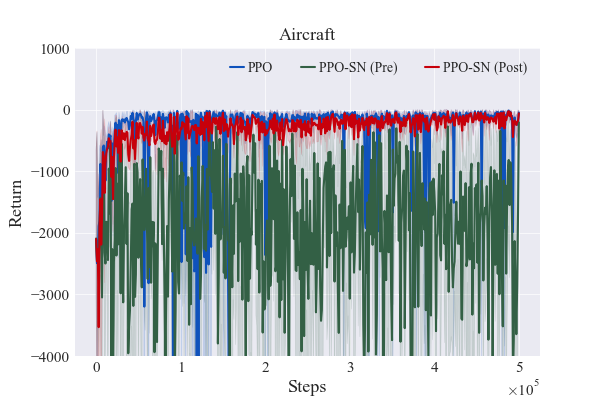}
          \hspace{1.6cm} (a) Learning curves.
        \end{center}
      \end{minipage}

      \begin{minipage}{0.5\hsize}
        \begin{center}
          \includegraphics[clip, width=8.3cm]{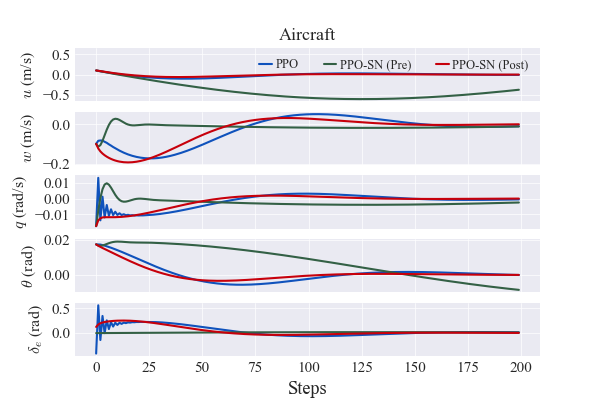}
          \hspace{1.6cm} (b) Trajectories.
        \end{center}
      \end{minipage} \\

      \begin{minipage}{0.5\hsize}
        \begin{center}
          \includegraphics[clip, width=8.2cm]{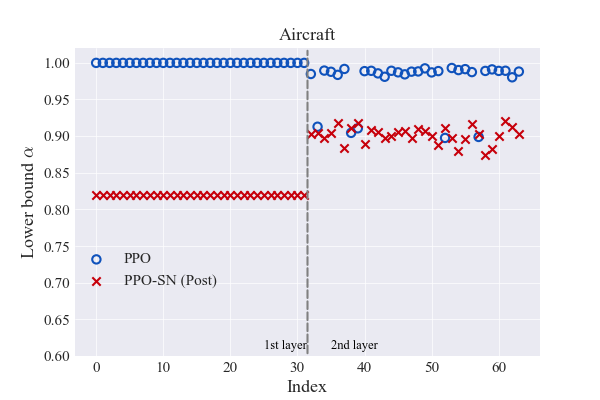}
          \hspace{2.6cm} (c) Calculated lower bounds.
        \end{center}
      \end{minipage}
      
      \begin{minipage}{0.5\hsize}
        \begin{center}
          \includegraphics[clip, width=8.2cm]{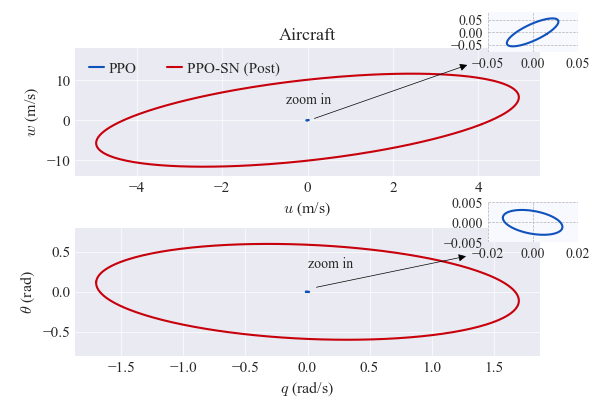}
          \hspace{2.6cm} (d) ROA.
        \end{center}
      \end{minipage}

    \end{tabular}
    \caption{Comparison between PPO and PPO-SN (Pre/Post) on the aircraft control task.}
    \label{Fig:ResultAircraft}
  \end{center}
\end{figure}

\subsection{Aircraft}
The aircraft model used in this article is the generic transport model (GTM) developed by NASA~\cite{jordan2004development}, whose nonlinear simulation model is available in~\cite{nasaGTM}. 
The linearized aircraft model is given by linearizing the nonlinear model at a trim point in accordance with~\cite{sawada2019worst}. 
For this experiments, the model is discretized via a zero-order hold at a sampling period of $T_s=0.05$ s. 
The $\mathcal{L}_2$ gain (i.e. $H_\infty$ norm) of the aircraft model is $\gamma_G=30.8$. 
Regarding the {\it pre-guaranteed RL}, the parameters on the spectral normalization are set to $\delta^i=0.31$, $i=1,...,3$, to satisfy the stability condition of the feedback system. 
The policy network has two hidden layers of 32 units.

Figure~\ref{Fig:ResultAircraft} shows the result of the aircraft control task.
The plot layout is the same as in Fig.~\ref{Fig:ResultPendulum} except for the plots (a) and (b) which include the result of the {\it pre-guaranteed RL}. 
Regarding the learning curve, the {\it pre-guaranteed RL} shows the insufficient performance due to the stability condition derived from the small-gain theorem. On the other hand, the {\it post-guaranteed RL} shows almost similar performance compared with the baseline PPO. Regarding the time history of the state as well, the {\it post-guaranteed RL} shows enough performance. Moreover, the ROA of the {\it post-guaranteed RL} is larger than that of the baseline PPO. 
These results show that the effectiveness of the spectral normalization for the stability-certified RL.

\begin{table}[!!t]
\begin{center}
\caption{Spectral norm of the weight matrices obtained on the aircraft control task. The spectral normalization is performed for all layers in PPO-SN (Pre) and only for hidden layers in PPO-SN (Post).}
\vspace{2mm}
\begin{tabular}{cccccc}\hline
& PPO & PPO-SN (Pre) & PPO-SN (Post) \\
&-&$\delta=0.31$ & $\delta=1.0$ \\\hline
$W_1$ & 9.598 & 0.3100 &  1.000  \\
$W_2$ & 9.360 & 0.3100 &  1.000  \\
$W_3$ & 2.411 & 0.3100 &  24.00  \\ \hline    
\end{tabular}
\label{tab:WeightComparisonAircraft}
\end{center}
\end{table}
\section{Conclusion}
\label{Sec:Conclusion}
In this article, to achieve a stability-certified RL, we have revisited the spectral normalization and proposed two types of methods from different perspectives. 
The first one is the {\it pre-guaranteed RL} to ensure the stability condition derived from the small-gain theorem. While explicitly including the stability condition in the training of the policy, the {\it pre-guaranteed RL} may provide insufficient performance due to the strict stability condition. In order to improve the performance while ensuring the stability, the {\it post-guaranteed RL} has been proposed, which much improves the feasibility of the a-posteriori stability analysis formulated by LMIs in many cases. 
The numerical experiments show that the {\it post-guaranteed RL} achieves almost similar performance compared with the baseline PPO while providing enough stability with a larger ROA. 
\section*{Acknowledgements}
The authors would like to thank Prof. Takashi Shimomura at Osaka Prefecture University for providing advice based on his expert knowledge of control theory.

\bibliographystyle{IEEEtran}
\bibliography{ms}

\appendix
\def\thesection{\Alph{section}}
\section{Environment Details}
\label{Sec:EnvDetails}
\subsection*{Inverted Pendulum}
The linearized equation of motion for the inverted pendulum is given as follows:
\begin{eqnarray}
A_G=I_2+T_s
\left[
\begin{array}{cc}
0&1\\
\frac{g}{l}&-\frac{\eta}{Ml^2}
\end{array}
\right], \quad
B_G = T_s
\left[
\begin{array}{cc}
0\\
\frac{1}{Ml^2}
\end{array}
\right]
\end{eqnarray}
where $l$, $M$, $g$, $\eta$, and $T_s$ are the length of the pendulum (m), the mass (kg), the gravitation constant ($\rm m/s^2$), the friction coefficient ($\rm Pa\cdot s$), and the sampling period (s), respectively. The parameters are set to the same as~\cite{okawa2019control} and summarized in Table~\ref{Tab:ParamsPendulum}. 
The state and the input are given by
\begin{eqnarray}
x(k)=\left[
\begin{array}{cccc}
\theta(k)&\dot{\theta}(k)
\end{array}
\right]^T, \quad u(k)=\tau(k)
\end{eqnarray}
where $\theta(k)$, $\dot{\theta}(k)$, and $\tau(k)$ are the angle (rad), the angular velocity (rad/s), and the torque ($\rm N\cdot m$), respectively. The system does not have a finite $\mathcal{L}_2$ gain. At each step, the reward is given by
\begin{eqnarray}
r(k) = -(x(k)^TQx(k)+u(k)^TRu(k))
\end{eqnarray}
where $Q={\rm diag}\{2,2\}$ and $R=0.001$. 
The current episode is terminated if $|\theta|>4\pi$ rad.
The max episode step is 200.

\begin{table}[htb]
  \begin{center}
    \caption{Parameters of the inverted pendulum.}
    \vspace{2mm}
    \begin{tabular}{cccc} \hline
      Symbol & Definition & Value \\ \hline
      $l$ & length of pendulum (m) & 0.5\\
      $M$& Mass (kg) & 0.15\\
      $g$ & gravitation constant ($\rm m/s^2$) & 9.8\\
      $\eta$ & Friction coefficient ($\rm Pa\cdot s$)&0.05\\
      $T_s$ & sampling period (s) & 0.1\\\hline
    \end{tabular}
    \label{Tab:ParamsPendulum}
  \end{center}
\end{table}

\subsection*{Aircraft}
The aircraft model is the generic transport model (GTM) developed by NASA, which is a dynamically scaled 5.5\% free-flying model of a twin-jet commercial transport aircraft~\cite{jordan2004development}. 
The nonlinear simulation model is available in~\cite{nasaGTM}. 
The linearized model of GTM is given by linearizing the nonlinear model at a trim point. 
In this article, the trim condition and the continuous-time model are taken from~\cite{sawada2019worst}. 
For the experiment of this article, the model is discretized via a zero-order hold at a sampling period of $T_s=0.05$ seconds. 
Thus, the discrete-time model for the aircraft longitudinal dynamics is given as follows.
\begin{eqnarray}
A_G=\left[
\begin{array}{rrrr}
9.968e^{-1}&1.530e^{-2}&-1.079e^{-1}&-4.891e^{-1}\\
-8.300e^{-3}&7.932e^{-1}&1.938\hspace{5.5mm}&-2.090e^{-2}\\
1.900e^{-3}&-5.050e^{-2}&7.436e^{-1}&2.405e^{-4}\\
4.681e^{-5}&-1.362e^{-3}&4.386e^{-2}&1.000\hspace{5.5mm}
\end{array}
\right], \quad
B_G=\left[
\begin{array}{rr}
3.145e^{-3}\\
-7.140e^{-2}\\
-4.969e^{-2}\\
-1.306e^{-3}
\end{array}
\right]
\end{eqnarray}
The state $x(k)$ and the input $u(k)$ are given as follows.
\begin{eqnarray}
x(k)=\left[
\begin{array}{cccc}
u(k)&w(k)&q(k)&\theta(k)
\end{array}
\right]^T, \quad u(k)=\delta_e(k)
\end{eqnarray}
where $u(k)$, $w(k)$, $q(k)$, $\theta(k)$, and $\delta_e(k)$ are the velocity perturbation in $X$- and $Z$-directions (m/s), the pitch rate (rad/s), the pitch angle (rad), and the elevator deflection (rad), respectively. 
The $\mathcal{L}_2$ gain (i.e. $H_\infty$ norm) is given by $\gamma_G=30.8$. At each step, the reward is given by
\begin{eqnarray}
r(k) = -(x(k)^TQx(k)+u(k)^TRu(k))
\end{eqnarray}
where $Q={\rm diag}\{2,2,2,2\}$ and $R=10$. 
The max episode step is 200. 
If $|\theta|>\pi/2$ rad, $|u|>5$ m/s, or $|w|>5$ m/s, the current episode is terminated.

\end{document}